\documentclass{article}

\usepackage{microtype}
\usepackage{graphicx}
\usepackage{booktabs}
\usepackage{amsmath}
\usepackage{subfig}
\usepackage{hyperref}

\newcommand{\Ignore}[1]{}
\usepackage[accepted]{icml2021}

\icmltitlerunning{Gradient-Based Interpretability Methods and Binarized Neural Networks}

\begin{document}

\twocolumn[
\icmltitle{Gradient-Based Interpretability Methods and Binarized Neural Networks}

\begin{icmlauthorlist}
\icmlauthor{Amy  Widdicombe}{ucl}
\icmlauthor{Simon J. Julier}{ucl}

\end{icmlauthorlist}

\icmlaffiliation{ucl}{Department of Computer Science, University College London, Gower Street, London, UK}

\icmlcorrespondingauthor{Amy Widdicombe}{amy.widdicombe.17@ucl.ac.uk}

\icmlkeywords{Machine Learning, ICML}

\vskip 0.3in
]

\printAffiliationsAndNotice{}

\begin{abstract}
Binarized Neural Networks (BNNs) have the potential to revolutionize the way that deep learning is carried out in edge computing platforms. However, the effectiveness of interpretability methods on these networks has not been assessed.

In this paper, we compare the performance of several widely used saliency map-based interpretabilty techniques (Gradient, SmoothGrad and GradCAM), when applied to Binarized or Full Precision Neural Networks (FPNNs).  We found that the basic Gradient method produces very similar-looking maps for both types of network. However, SmoothGrad produces significantly noisier maps for BNNs. GradCAM also produces saliency maps which differ between network types, with some of the BNNs having seemingly nonsensical explanations. We comment on possible reasons for these differences in explanations and present it as an example of why interpretability techniques should be tested on a wider range of network types.
\end{abstract}

\section{Introduction}\label{introduction}
Binarized Neural networks (BNNs) \cite{DBLP:journals/corr/CourbariauxB16} have been proposed as a way to create small, power and memory efficient networks which could, with specialist hardware, provide a more efficient alternative to existing full precision networks (FPNNs). When choosing an efficient network to deploy to an edge device, many properties will be considered; size, accuracy, efficiency and, particularly in safety critical circumstances, adversarial robustness and interpretability. Although studies have been carried out to assess the robustness of BNNs to adversarial attacks \cite{galloway2018attacking, lin2018defensive}, as far as we are aware no similar studies have been carried out into the interpretability of these networks.

In this workshop paper, we begin an exploration of this critical gap in the assessment of BNNs. While ``interpretable'' can mean many things, here we focus on gradient-based saliency methods. These methods have been extensively investigated and are widely used.  We compare the effectiveness of these methods when applied to BNNs and FPNNs. 

We investigate three gradient-based interpretability techniques Gradient \cite{DBLP:journals/corr/SimonyanVZ13}, SmoothGrad \cite{DBLP:journals/corr/SmilkovTKVW17} and GradCAM \cite{Selvaraju2019GradCAMVE}). Our results show that the Gradient method is largely the same for both network types. However, SmoothGrad and GradCAM produce visibly different results which, in some cases, do not appear to make sense.

\section{Related Work}\label{related}

\subsection{Binarized Neural Networks}\label{bnns}
Binarized Neural Networks (BNNs) \cite{DBLP:journals/corr/CourbariauxB16} are neural networks where the weights and activations are binary (usually $+1$ or $-1$). As such these different neural networks are intended to be more power and memory efficient and could be extremely useful in situations where hardware size and battery life are important. This increased efficiency can be achieved with minimal loss in accuracy; when initially proposed, BNNs were shown to produce results close to the state of the art on the MNIST, CIFAR-10 and SVHN data sets \cite{DBLP:journals/corr/CourbariauxB16} and since then different architectures have been proposed to improve BNN accuracy on more complex data sets such as ImageNet \cite{DBLP:journals/corr/abs-1711-11294}.

Binarization is achieved using the sign function and, in order to get around the non-differentiable nature of this function, pseudo-gradients are used for backward passes through the network \cite{DBLP:journals/corr/CourbariauxB16}. The most common of these is the ``straight-through estimator'' \cite{coursera/hinton}. The use of the sign function and pseudo gradients are reasons we might expect gradient based saliency methods to perform differently for BNNs. 

\subsection{Interpretability Techniques - Saliency Methods}\label{interp_tech}
Saliency methods are a type of interpretability technique which aim to highlight the parts of an input which are most important to a network's classification decision. While many such techniques exist, their reliability has been questioned. For example, Adebayo et al. use a series of ``sanity checks" to show that some techniques are invariant under parameter and data randomisation~\cite{DBLP:journals/corr/abs-1810-03292}. As we are comparing different types of networks, they must be sensitive to parameter values. Therefore, we chose three techniques which have passed these ``sanity checks": Gradient, SmoothGrad and GradCAM:

\textbf{Gradient:} One of the oldest methods for computing saliency maps, Gradient uses a single backward pass through the network, taking the derivative of the class score with respect to the input image, to highlight the most important pixels \cite{DBLP:journals/corr/SimonyanVZ13}.

\textbf{SmoothGrad:} SmoothGrad \cite{DBLP:journals/corr/SmilkovTKVW17} is a simple technique which can be bolted on to other saliency methods by adding a further step to the technique: by generating several input images perturbed with random Gaussian noise and taking the average of the resulting saliency maps, one is left with a less noisy final map.

\textbf{GradCAM:} GradCAM \cite{Selvaraju2019GradCAMVE} is a frequently used technique which uses the gradient of a class flowing into a network's last conv layer to produce its saliency maps. The maps produced are coarser than those created by methods which attribute to each pixel of an image; GradCAM instead highlights regions of an image, this is due to the fact that it upsamples the heatmap produced at the last conv layer to match the dimensions of the input image.

\section{Experiments}\label{experiments}

\subsection{Method and Networks Used}\label{networks_used}
We use the \texttt{tf-keras-vis toolkit}~\cite{tf-keras-vis} for our saliency methods. To compare the effect of binarization, we compute the performance of 5 types of CNNs and 5 types of BNNs, all of which have all been pre-trained on ImageNet \cite{deng2009imagenet}.

For full precision networks, we used the following implementations from \texttt{tf.keras.applications} \cite{tensorflow2015-whitepaper}: EfficientNet (the smallest, most efficient version --- B0) \cite{DBLP:conf/icml/TanL19}, MobileNet \cite{Howard2017MobileNetsEC} and MobileNetV2 \cite{Sandler2018MobileNetV2IR} because they are designed to be efficient and deployed in similar settings to those which BNNs target. We also use VGG16 \cite{Simonyan15} which is frequently used in interpretability technique experiments and ResNet50 \cite{He2016DeepRL}. 

Pre-trained BNNs are much harder to come by, so we choose from those found in the Larq Zoo library \cite{larq}, using one of the library's own state of the art pre-trained models, QuickNet \cite{Bannink2020LarqCE}, and four models from the literature; BinaryAlexNet \cite{NIPS2016_d8330f85}, BinaryResNetE18 \cite{DBLP:journals/corr/abs-1906-08637}, MeliusNet22 \cite{Bethge2020MeliusNetCB} and RealtoBinaryNet \cite{Martinez2020Training}. 

\subsection{Results --- Gradient}\label{grad}

\begin{figure}
\begin{center}
\subfloat[Full Precision Networks\label{fig:grad_bnns}]
{
\includegraphics[width=0.8\linewidth]
{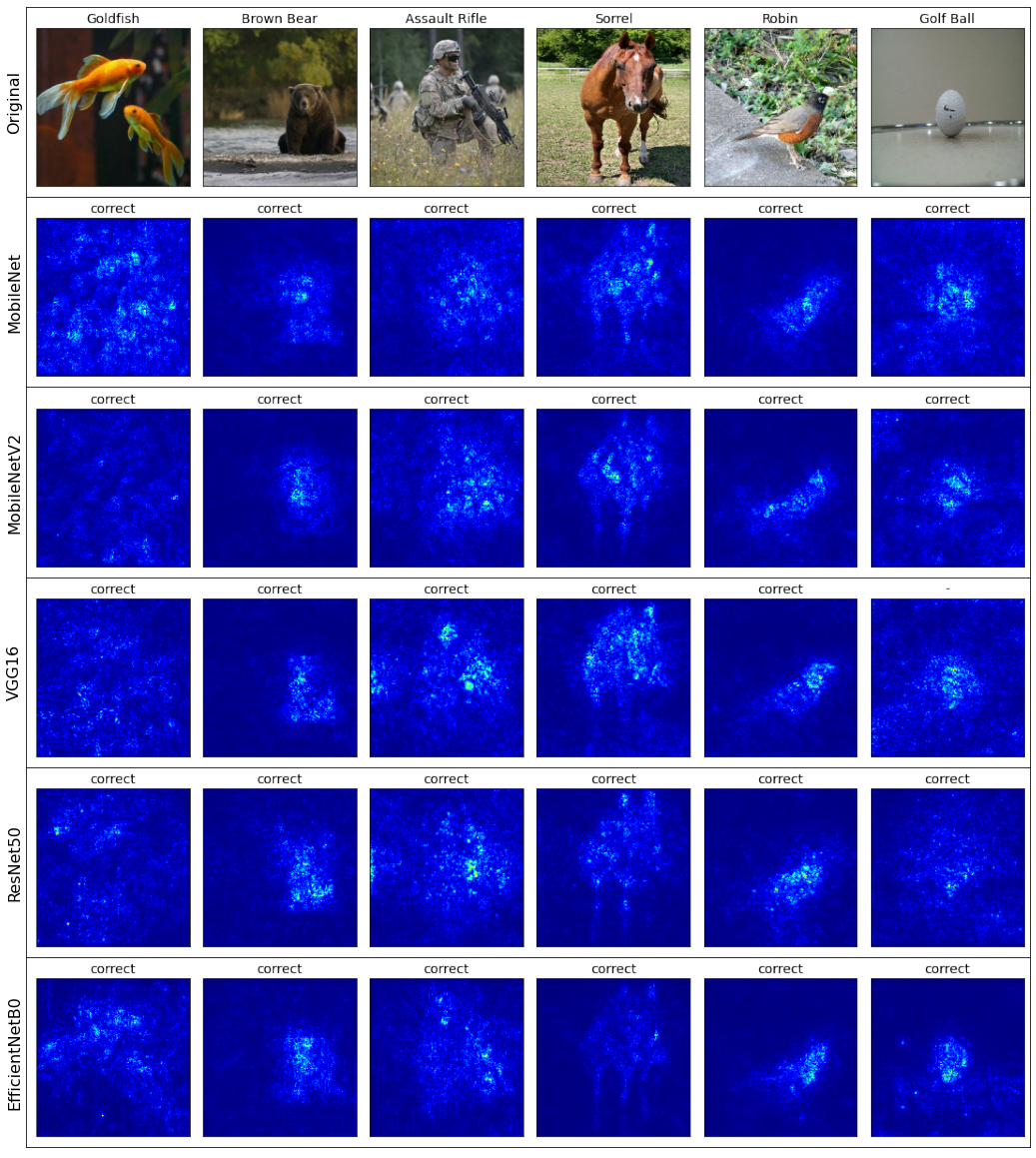}
}\\
\subfloat[Binarized Networks\label{fig:grad_fp}]{\includegraphics[width=0.8\linewidth]{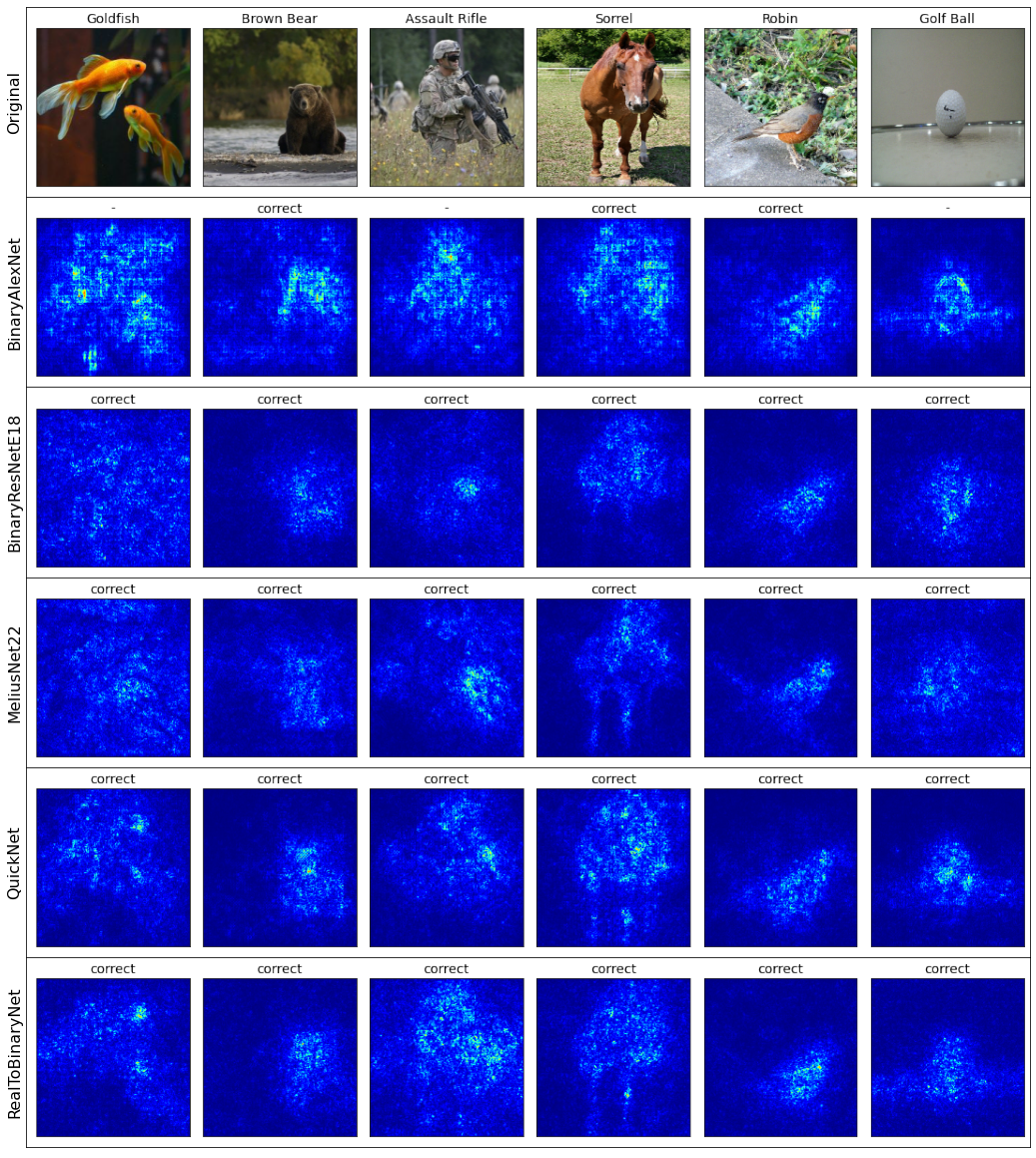}}
\end{center}
\caption{Comparison between gradient saliency maps for our 5 FPNNs and BNNs. If the network classified the image correctly, the saliency map has ``correct" written above it. The interpretability method seems to be working as expected and there is no obvious difference between the maps from the FPNNs and the maps from the BNNs.}
\label{fig:grad_exp}
\end{figure}

From Figure~\ref{fig:grad_exp}, the basic gradient technique seems to work as expected for BNNs. By visual comparison it is difficult to make distinction between the two different types of networks. This is evidence that the use of pseudo gradients (by BNNs) does not impact the visual quality of basic Gradient saliency maps. 

\subsection{Results --- SmoothGrad}\label{smoothgrad}

The original paper found that 10 -- 20\% noise is optimal and we start by applying 20\%. At this level of noise we can see in Figure~\ref{fig:grad_exp} that the BNNs produce much less clear saliency maps than the FPNNs --- rather than making the saliency maps less noisy, some of the maps appear more noisy than then plain gradient maps in the previous section. 

\begin{figure}
\vspace*{-1cm}
\begin{center}
\subfloat[Full Precision Networks\label{fig:smoothgrad_fp}]
{
\includegraphics[width=0.8\linewidth]
{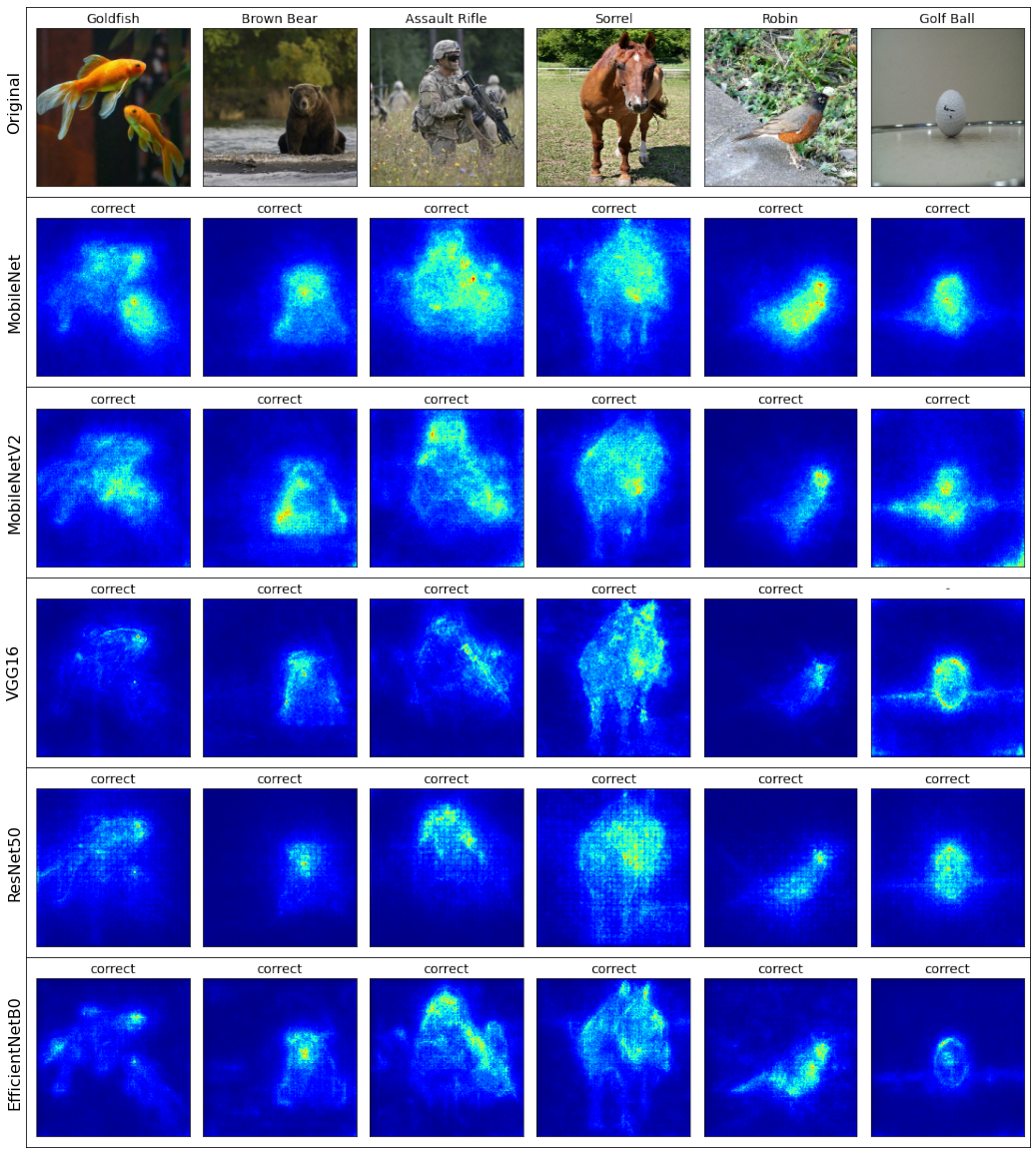}
}\\
\subfloat[Binarized Networks\label{fig:smoothgrad_bnns}]{\includegraphics[width=0.8\linewidth]{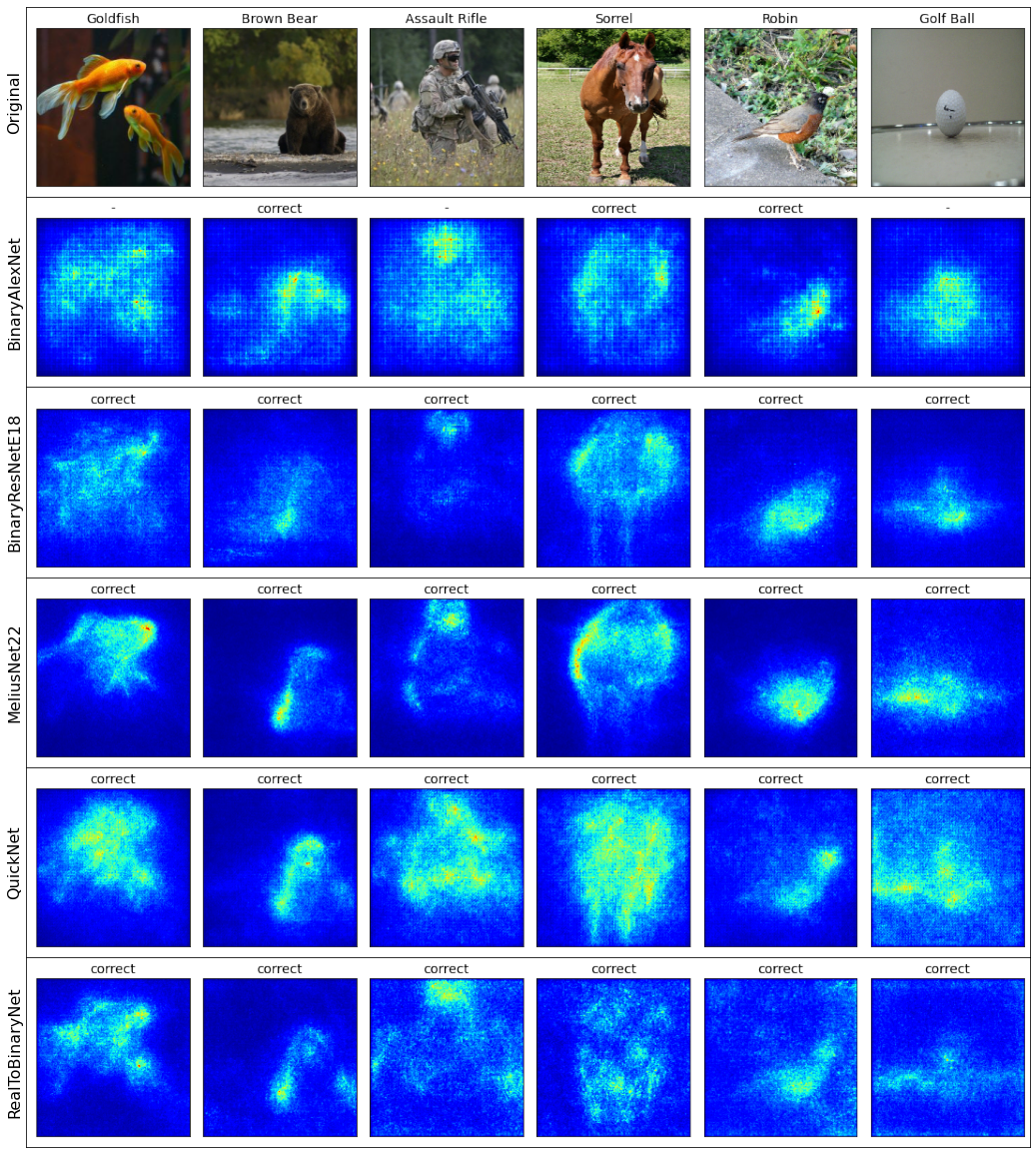}}
\end{center}
\caption{Comparison between SmoothGrad saliency maps for our 5 FPNNs and BNNs. As you can see, across all of the images, the BNN's saliency maps appear to be both less sharp and more noisy. }
\label{fig:smooth_grad_exp}
\end{figure}

We begin further investigations, starting with looking at maps produced across a range of different levels of noise perturbation, from 1\% to 50\%. We find that BNN's saliency maps degrade to being almost completely to random, covering most of the image by 50\% noise while the FPNNs retain at the very least some vague shape of the object. 

Although it varies from network to network, the optimal noise level for clarity of saliency map seems to be much lower for BNNs, around 2--6\%. A potential explanation for this is increased noise sensitivity. In \cite{lin2018defensive}, Lin et. al. analyzed the propagation of adversarial perturbations through quantized networks (including BNNs). They show that perturbations are amplified as they filter through each layer of a network and that quantization operations further amplify the perturbation. As a result, the more quantized a network is, the more the perturbations are amplified. 

Although Lin et. al look at adversarial attacks, a similar effect could appear with any form of perturbations, including the random sampling from SmoothGrad. Our experiments support this hypothesis: the need for a smaller perturbation in SmoothGrad implies that BNNs are more sensitive to noise than most FPNNs. 

\subsection{Results --- GradCAM}\label{GradCAM}

\begin{figure}
\begin{center}
\subfloat[Full Precision Networks\label{fig:GradCAM_fp}]
{
\includegraphics[width=0.8\linewidth]
{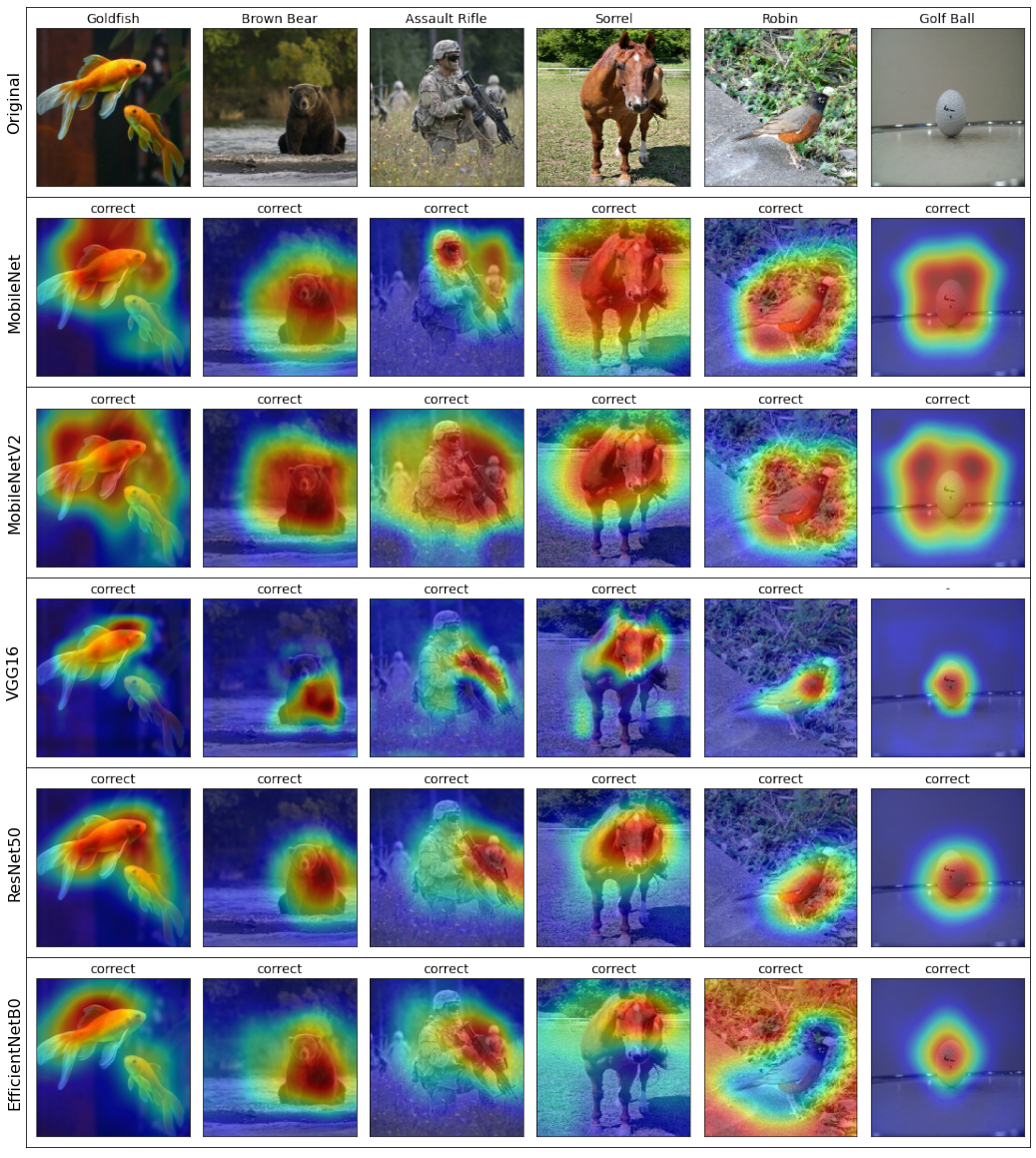}
}\\
\subfloat[Binarized Networks\label{fig:GradCam_bnns}]{\includegraphics[width=0.8\linewidth]{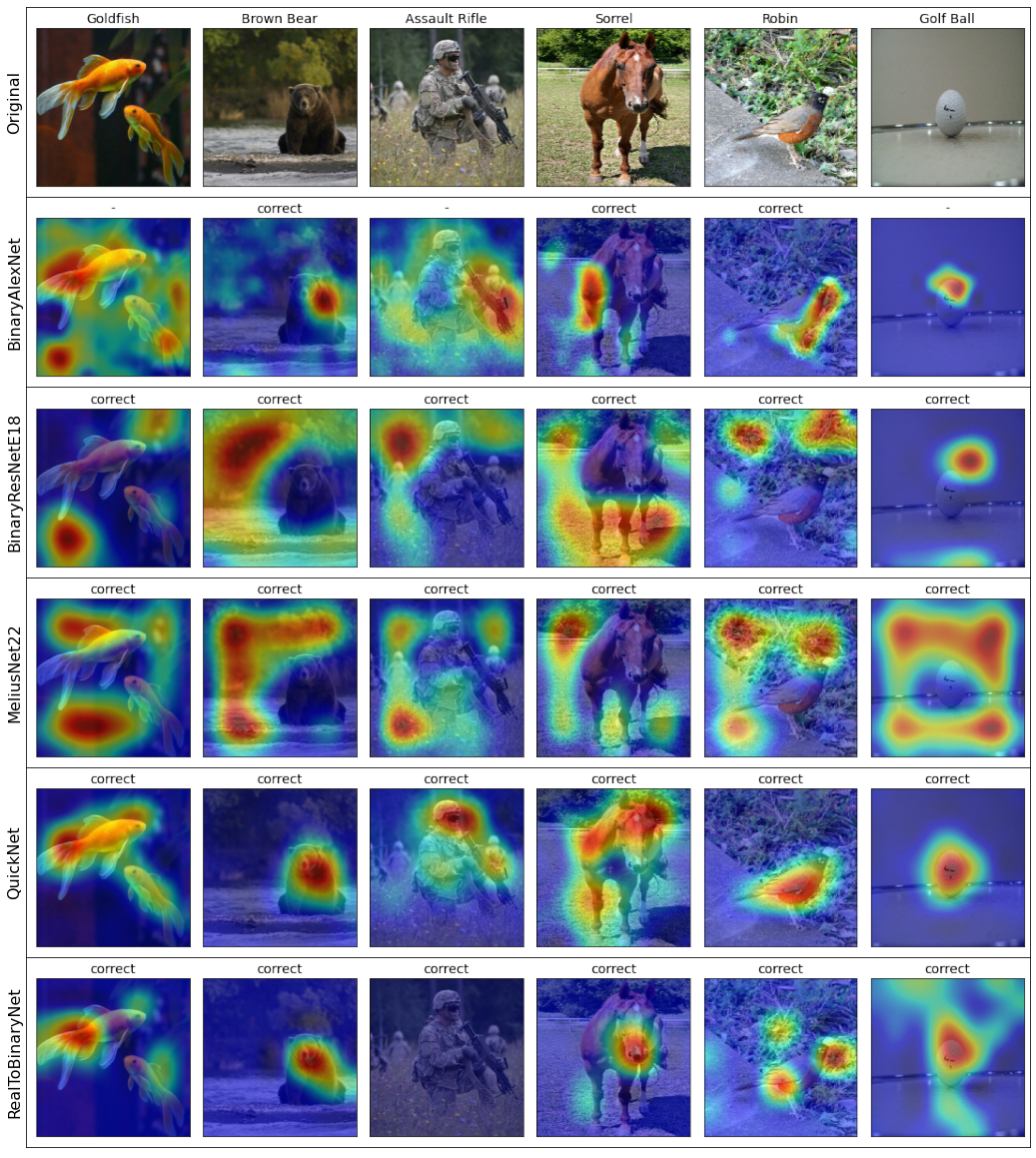}}
\end{center}
\caption{Results of applying GradCAM to our networks. Most of the full precision networks produce clear saliency maps highlighting the object of the class, although for the EfficientNetB0 map for the Robin image the attributions appear reversed. For the BNNs, very few of the maps appear reasonable, with only QuickNet consistently outputing maps which highlight the object as expected. One of the networks even returns a blank map for an image and for many of the others the attributions fall in strange parts of the image e.g. mostly on the background.}
\label{fig:GradCAM_exp}
\end{figure}

GradCAM is designed and expected to work for any CNN architecture \cite{Selvaraju2019GradCAMVE}, so our results in Figure~\ref{fig:GradCAM_exp} are surprising, particularly in the example where the method returns a completely blank saliency map and where seemingly uninformative and random seeming patches of the images are highlighted. 

\cite{Selvaraju2019GradCAMVE} justify GradCAM by claiming that the best balance between spatial information and high-level semantics is likely to be found in the last convolutional layer. It is possible that BNNs store spatial and semantic/class sensitive information in a different way to FPNNs and this causes the strange attribution patterns. It is also possible that (for BNNs) the ReLU in GradCAM is filtering out informative values --- if we remove the ReLU and invert the saliency map, some of the networks return maps closer to those from FPNNs.

\section{Conclusion and Future Work}\label{conclusion}

We have looked at how three gradient-based saliency methods work when applied to BNNs. We found that both GradCAM and SmoothGrad appear to behave differently than expected for these networks. For SmoothGrad this appears to be due to the increased noise sensitivity of BNNs and can be adjusted for by reducing the noise perturbation level. For GradCAM the explanation is less clear and, as not all of the BNNs exhibit the same behaviour, we cannot be certain which element of the network is responsible for these differences. Although these experiments are not conclusive, they provide an example of where well known interpretability methods behave differently than expected, or even fail. We hope that future work designing new interpretability techniques will consider this and apply these techniques to a wider range of networks for testing. Additionally, when new types of network are designed and tested, we think interpretability (by as many techniques as possible) should be considered and tested. 

Future work building on this could look at a wider range of interpretability techniques and network-types - as well attempting to isolate exactly which parts of a network design/architecture cause changes in the behaviour of methods like GradCAM. What, if anything, can interpretability techniques tell us about how differently two distinct network types are working? And what can this tell us about the interpretability techniques themselves?

\section*{Acknowledgements}

This research was sponsored by the U.S. Army Research Laboratory and the U.K. Ministry of Defence under Agreement Number W911NF-16-3-0001. The views and conclusions contained in this document are those of the authors and should not be interpreted as representing the official policies, either expressed or implied, of the U.S. Army Research Laboratory, the U.S. Government, the U.K. Ministry of Defence or the U.K. Government. The U.S. and U.K. Governments are authorized to reproduce and distribute reprints for Government purposes notwithstanding any copyright notation hereon.

\bibliography{biblio_2021}

\begin{thebibliography}{23}
\providecommand{\natexlab}[1]{#1}
\providecommand{\url}[1]{\texttt{#1}}
\expandafter\ifx\csname urlstyle\endcsname\relax
  \providecommand{\doi}[1]{doi: #1}\else
  \providecommand{\doi}{doi: \begingroup \urlstyle{rm}\Url}\fi

\bibitem[Abadi et~al.(2015)Abadi, Agarwal, Barham, Brevdo, Chen, Citro,
  Corrado, Davis, Dean, Devin, Ghemawat, Goodfellow, Harp, Irving, Isard, Jia,
  Jozefowicz, Kaiser, Kudlur, Levenberg, Man\'{e}, Monga, Moore, Murray, Olah,
  Schuster, Shlens, Steiner, Sutskever, Talwar, Tucker, Vanhoucke, Vasudevan,
  Vi\'{e}gas, Vinyals, Warden, Wattenberg, Wicke, Yu, and
  Zheng]{tensorflow2015-whitepaper}
Abadi, M., Agarwal, A., Barham, P., Brevdo, E., Chen, Z., Citro, C., Corrado,
  G.~S., Davis, A., Dean, J., Devin, M., Ghemawat, S., Goodfellow, I., Harp,
  A., Irving, G., Isard, M., Jia, Y., Jozefowicz, R., Kaiser, L., Kudlur, M.,
  Levenberg, J., Man\'{e}, D., Monga, R., Moore, S., Murray, D., Olah, C.,
  Schuster, M., Shlens, J., Steiner, B., Sutskever, I., Talwar, K., Tucker, P.,
  Vanhoucke, V., Vasudevan, V., Vi\'{e}gas, F., Vinyals, O., Warden, P.,
  Wattenberg, M., Wicke, M., Yu, Y., and Zheng, X.
\newblock {TensorFlow}: Large-scale machine learning on heterogeneous systems,
  2015.
\newblock URL \url{https://www.tensorflow.org/}.
\newblock Software available from tensorflow.org.

\bibitem[Adebayo et~al.(2018)Adebayo, Gilmer, Muelly, Goodfellow, Hardt, and
  Kim]{DBLP:journals/corr/abs-1810-03292}
Adebayo, J., Gilmer, J., Muelly, M., Goodfellow, I.~J., Hardt, M., and Kim, B.
\newblock Sanity checks for saliency maps.
\newblock \emph{CoRR}, abs/1810.03292, 2018.
\newblock URL \url{http://arxiv.org/abs/1810.03292}.

\bibitem[Bannink et~al.(2020)Bannink, Hillier, Geiger, Bruin, Overweel, Neeven,
  and Helwegen]{Bannink2020LarqCE}
Bannink, T., Hillier, A.~C., Geiger, L., Bruin, T.~D., Overweel, L., Neeven,
  J., and Helwegen, K.
\newblock Larq compute engine: Design, benchmark, and deploy state-of-the-art
  binarized neural networks.
\newblock \emph{ArXiv}, abs/2011.09398, 2020.

\bibitem[Bethge et~al.(2019)Bethge, Yang, Bornstein, and
  Meinel]{DBLP:journals/corr/abs-1906-08637}
Bethge, J., Yang, H., Bornstein, M., and Meinel, C.
\newblock Back to simplicity: How to train accurate bnns from scratch?
\newblock \emph{CoRR}, abs/1906.08637, 2019.
\newblock URL \url{http://arxiv.org/abs/1906.08637}.

\bibitem[Bethge et~al.(2020)Bethge, Bartz, Yang, Chen, and
  Meinel]{Bethge2020MeliusNetCB}
Bethge, J., Bartz, C., Yang, H., Chen, Y., and Meinel, C.
\newblock Meliusnet: Can binary neural networks achieve mobilenet-level
  accuracy?
\newblock \emph{ArXiv}, abs/2001.05936, 2020.

\bibitem[Courbariaux \& Bengio(2016)Courbariaux and
  Bengio]{DBLP:journals/corr/CourbariauxB16}
Courbariaux, M. and Bengio, Y.
\newblock Binarynet: Training deep neural networks with weights and activations
  constrained to +1 or -1.
\newblock \emph{CoRR}, abs/1602.02830, 2016.
\newblock URL \url{http://arxiv.org/abs/1602.02830}.

\bibitem[Deng et~al.(2009)Deng, Dong, Socher, Li, Li, and
  Fei-Fei]{deng2009imagenet}
Deng, J., Dong, W., Socher, R., Li, L.-J., Li, K., and Fei-Fei, L.
\newblock Imagenet: A large-scale hierarchical image database.
\newblock In \emph{2009 IEEE conference on computer vision and pattern
  recognition}, pp.\  248--255. Ieee, 2009.

\bibitem[Galloway et~al.(2018)Galloway, Taylor, and
  Moussa]{galloway2018attacking}
Galloway, A., Taylor, G.~W., and Moussa, M.
\newblock Attacking binarized neural networks.
\newblock In \emph{International Conference on Learning Representations}, 2018.
\newblock URL \url{https://openreview.net/forum?id=HkTEFfZRb}.

\bibitem[Geiger \& Team(2020)Geiger and Team]{larq}
Geiger, L. and Team, P.
\newblock Larq: An open-source library for training binarized neural networks.
\newblock \emph{Journal of Open Source Software}, 5\penalty0 (45):\penalty0
  1746, January 2020.
\newblock \doi{10.21105/joss.01746}.
\newblock URL \url{https://doi.org/10.21105/joss.01746}.

\bibitem[He et~al.(2016)He, Zhang, Ren, and Sun]{He2016DeepRL}
He, K., Zhang, X., Ren, S., and Sun, J.
\newblock Deep residual learning for image recognition.
\newblock \emph{2016 IEEE Conference on Computer Vision and Pattern Recognition
  (CVPR)}, pp.\  770--778, 2016.

\bibitem[Hinton(2012)]{coursera/hinton}
Hinton, G.
\newblock Neural networks for machine learning.
\newblock "Coursera, video lectures", 2012.

\bibitem[Howard et~al.(2017)Howard, Zhu, Chen, Kalenichenko, Wang, Weyand,
  Andreetto, and Adam]{Howard2017MobileNetsEC}
Howard, A., Zhu, M., Chen, B., Kalenichenko, D., Wang, W., Weyand, T.,
  Andreetto, M., and Adam, H.
\newblock Mobilenets: Efficient convolutional neural networks for mobile vision
  applications.
\newblock \emph{ArXiv}, abs/1704.04861, 2017.

\bibitem[Hubara et~al.(2016)Hubara, Courbariaux, Soudry, El-Yaniv, and
  Bengio]{NIPS2016_d8330f85}
Hubara, I., Courbariaux, M., Soudry, D., El-Yaniv, R., and Bengio, Y.
\newblock Binarized neural networks.
\newblock In Lee, D., Sugiyama, M., Luxburg, U., Guyon, I., and Garnett, R.
  (eds.), \emph{Advances in Neural Information Processing Systems}, volume~29,
  pp.\  4107--4115. Curran Associates, Inc., 2016.
\newblock URL
  \url{https://proceedings.neurips.cc/paper/2016/file/d8330f857a17c53d217014ee776bfd50-Paper.pdf}.

\bibitem[Kubota()]{tf-keras-vis}
Kubota, Y.
\newblock tf-keras-vis.
\newblock URL \url{https://github.com/keisen/tf-keras-vis}.
\newblock visualization toolkit for debugging tf.keras models in
  Tensorflow2.0+.

\bibitem[Lin et~al.(2019)Lin, Gan, and Han]{lin2018defensive}
Lin, J., Gan, C., and Han, S.
\newblock Defensive quantization: When efficiency meets robustness.
\newblock In \emph{International Conference on Learning Representations}, 2019.
\newblock URL \url{https://openreview.net/forum?id=ryetZ20ctX}.

\bibitem[Lin et~al.(2017)Lin, Zhao, and Pan]{DBLP:journals/corr/abs-1711-11294}
Lin, X., Zhao, C., and Pan, W.
\newblock Towards accurate binary convolutional neural network.
\newblock \emph{CoRR}, abs/1711.11294, 2017.
\newblock URL \url{http://arxiv.org/abs/1711.11294}.

\bibitem[Martinez et~al.(2020)Martinez, Yang, Bulat, and
  Tzimiropoulos]{Martinez2020Training}
Martinez, B., Yang, J., Bulat, A., and Tzimiropoulos, G.
\newblock Training binary neural networks with real-to-binary convolutions.
\newblock In \emph{International Conference on Learning Representations}, 2020.
\newblock URL \url{https://openreview.net/forum?id=BJg4NgBKvH}.

\bibitem[Sandler et~al.(2018)Sandler, Howard, Zhu, Zhmoginov, and
  Chen]{Sandler2018MobileNetV2IR}
Sandler, M., Howard, A., Zhu, M., Zhmoginov, A., and Chen, L.-C.
\newblock Mobilenetv2: Inverted residuals and linear bottlenecks.
\newblock \emph{2018 IEEE/CVF Conference on Computer Vision and Pattern
  Recognition}, pp.\  4510--4520, 2018.

\bibitem[Selvaraju et~al.(2019)Selvaraju, Das, Vedantam, Cogswell, Parikh, and
  Batra]{Selvaraju2019GradCAMVE}
Selvaraju, R.~R., Das, A., Vedantam, R., Cogswell, M., Parikh, D., and Batra,
  D.
\newblock Grad-cam: Visual explanations from deep networks via gradient-based
  localization.
\newblock \emph{International Journal of Computer Vision}, 128:\penalty0
  336--359, 2019.

\bibitem[Simonyan \& Zisserman(2015)Simonyan and Zisserman]{Simonyan15}
Simonyan, K. and Zisserman, A.
\newblock Very deep convolutional networks for large-scale image recognition.
\newblock In \emph{International Conference on Learning Representations}, 2015.

\bibitem[Simonyan et~al.(2013)Simonyan, Vedaldi, and
  Zisserman]{DBLP:journals/corr/SimonyanVZ13}
Simonyan, K., Vedaldi, A., and Zisserman, A.
\newblock Deep inside convolutional networks: Visualising image classification
  models and saliency maps.
\newblock \emph{CoRR}, abs/1312.6034, 2013.
\newblock URL \url{http://arxiv.org/abs/1312.6034}.

\bibitem[Smilkov et~al.(2017)Smilkov, Thorat, Kim, Vi{\'{e}}gas, and
  Wattenberg]{DBLP:journals/corr/SmilkovTKVW17}
Smilkov, D., Thorat, N., Kim, B., Vi{\'{e}}gas, F.~B., and Wattenberg, M.
\newblock Smoothgrad: removing noise by adding noise.
\newblock \emph{CoRR}, abs/1706.03825, 2017.
\newblock URL \url{http://arxiv.org/abs/1706.03825}.

\bibitem[Tan \& Le(2019)Tan and Le]{DBLP:conf/icml/TanL19}
Tan, M. and Le, Q.~V.
\newblock Efficientnet: Rethinking model scaling for convolutional neural
  networks.
\newblock In \emph{ICML}, pp.\  6105--6114, 2019.
\newblock URL \url{http://proceedings.mlr.press/v97/tan19a.html}.

\end{thebibliography}
\bibliographystyle{icml2021}

\end{document}